\documentclass[10pt,twocolumn,letterpaper]{article}

\usepackage{cvpr}
\usepackage{times}
\usepackage{epsfig}
\usepackage{graphicx}
\usepackage{amsmath}
\usepackage{amssymb}
\usepackage{multirow}
\graphicspath{{eps/}{img/}}

\usepackage[breaklinks=true,bookmarks=false]{hyperref}

\cvprfinalcopy 

\def\cvprPaperID{1} 

\ifcvprfinal\fi
\begin{document}

\title{Dedicated Inference Engine and Binary-Weight Neural Networks for Lightweight Instance Segmentation}
\author{Tse-Wei Chen$^*$,
~Wei Tao$^\dag$,
~Dongyue Zhao$^\dag$,
~Kazuhiro Mima$^*$,\\
~Tadayuki Ito$^*$,
~Kinya Osa$^*$,
~and Masami Kato$^*$\\
{\tt\small twchen@ieee.org}\\
\and
$^*$Device Technology Development Headquarters, Canon Inc.,\\
30-2, Shimomaruko 3-chome, Ohta-ku, Tokyo 146-8501, Japan\\
$^\dag$Canon Innovative Solution (Beijing) Co., Ltd.,\\
12A Floor, Yingu Building, No.9 Beisihuanxi Road, Haidian, Beijing, China\\
}


\maketitle
\thispagestyle{empty}

\maketitle

\begin{abstract}
Reducing computational costs is an important issue for development of embedded systems. Binary-weight Neural Networks (BNNs), in which weights are binarized and activations are quantized, are employed to reduce computational costs of various kinds of applications. In this paper, a design methodology of hardware architecture for inference engines is proposed to handle modern BNNs with two operation modes. Multiply-Accumulate (MAC) operations can be simplified by replacing multiply operations with bitwise operations. The proposed method can effectively reduce the gate count of inference engines by removing a part of computational costs from the hardware system. The architecture of MAC operations can calculate the inference results of BNNs efficiently with only 52\% of hardware costs compared with the related works. To show that the inference engine can handle practical applications, two lightweight networks which combine the backbones of SegNeXt and the decoder of SparseInst for instance segmentation are also proposed. The output results of the lightweight networks are computed using only bitwise operations and add operations. The proposed inference engine has lower hardware costs than related works. The experimental results show that the proposed inference engine can handle the proposed instance-segmentation networks and achieves higher accuracy than YOLACT on the ``Person" category although the model size is 77.7$\times$ smaller compared with YOLACT. 
\end{abstract}

\section{Introduction}

Deep neural networks, including convolutional neural networks (CNN) and vision transformers (ViT)~\cite{Ranftl21}, have received considerable attention in recent years. Various kinds of networks have already been applied to different computer vision tasks. Not only can these algorithms be applied to face detection and object detection~\cite{Deng19,Terven23}, they can also be employed in other dense-prediction tasks such as semantic segmentation~\cite{Csurka23} and instance segmentation~\cite{Bolya19,Cheng22}.

For mobile devices and embedded systems with limited computational resources, it is necessary to reduce computational costs and power consumption. In modern neural networks, Multiply-Accumulate (MAC) operations have a much higher ratio than any other operations, such as max-pooling or up-sampling operations. Many kinds of quantization algorithms~\cite{Choi18,Sambhav20} and low-bit networks~\cite{Courbariaux15,Rastegari16} are proposed to simplify the MAC operations, and many kinds of hardware architectures are proposed to handle mixed-precision networks with low bit widths~\cite{Agrawal21} and binary-weight neural networks (BNNs) with low computational costs~\cite{Chen21,Chung23,Guo23,Guo19,Hu18,Tsai20,Yonekawa17}. However, the accuracy might decrease and fail to satisfy the requirement of some applications, such as instance segmentation, when the weights of networks are binarized or quantized to low bits. To design a suitable system for embedded vision, it is necessary to seek the balance between accuracy and bit widths in the quantized networks.

In this paper, we focus on both hardware implementation methods and algorithm design approaches of BNNs, in which activations are quantized and weights are binarized. Two lightweight networks for instance segmentation and a new hardware architecture of inference engine are proposed. The contribution of this paper is twofold. First, we propose a new design methodology to reduce the gate count of the inference engine by removing a part of computational costs from the hardware system. The methodology can be applied to different variations of BNNs, and no multiply operations are required. The values of binary weights can be either $\{0, 1\}$ or $\pm 1$. Second, we show that the proposed inference engine can handle instance segmentation algorithms with only 9.8\% of computational costs compared with the related works. The proposed algorithm and hardware can handle dense-prediction tasks with acceptable accuracy.

The paper is organized as follows. In Sec.~\ref{sec:related}, the related works are introduced. The architecture of the proposed inference engine for BNNs is shown in Sec.~\ref{sec:proposed}. The experimental results are discussed in Sec.~\ref{sec:experiments}. The conclusions are given in Sec.~\ref{sec:conclusions}.

\begin{figure*}[ht]
\begin{center}
\begin{tabular}{cc}
   \includegraphics[width=0.95\linewidth]{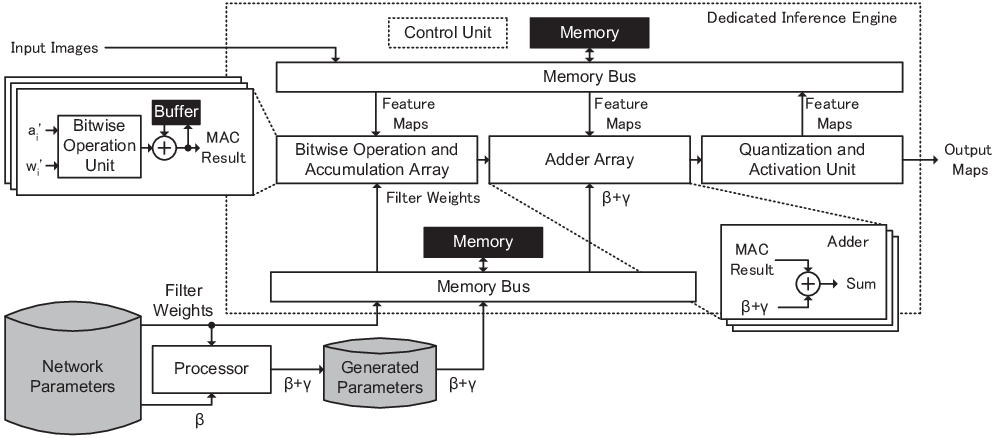}\\
\end{tabular}
\end{center}
   \caption{Hardware architecture of the proposed dedicated inference engine.}
\label{fig:architecture-a}
\end{figure*}

\begin{figure}[ht]
\begin{center}
\begin{tabular}{c}
   \includegraphics[height=0.9\linewidth]{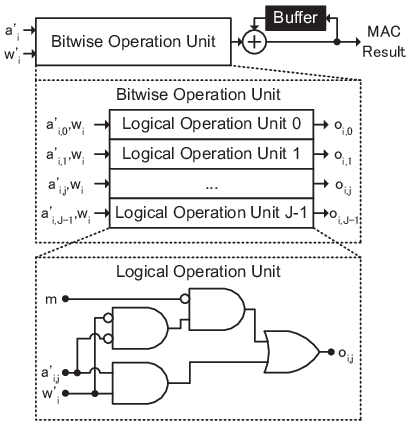}\\
\end{tabular}
\end{center}
   \caption{Hardware architecture of the ``bitwise operation unit" in the dedicated inference engine.}
\label{fig:architecture-b}
\end{figure}

\section{Binary-Weight Neural Networks (BNNs)}
\label{sec:related}

Quantization is an important topic for hardware implementation of neural networks. Many algorithms are proposed to quantize both activations and weights of CNN. Choi et al.~\cite{Choi18} propose a quantization scheme for activations during training. Sambhav et al.~\cite{Sambhav20} propose a method of training quantization thresholds, where the quantizers are suitable for hardware implementation. Gao et al.~\cite{Gao19} propose a systematic approach to transform the parameters of low-bit quantized networks into multiple thresholds for hardware implementation.

A number of researchers have implemented diverse approaches to design BNNs~\cite{Qin20}. BinaryConnect is a method which focuses on training networks with binary weights during forward and backward propagations~\cite{Courbariaux15}. To further reduce the computational costs, weights or feature maps (activations) can be quantized to 1 bit. XNOR-Net~\cite{Rastegari16} is a network where weights and activations are binarized, so that add operations can be replaced by logical operations, such as Exclusive-OR (XOR)~\cite{Zhu20} and Exclusive-NOR (XNOR) operations. However, when BNNs are employed in some practical applications, the accuracy decreases because the features extracted by the networks with binary weights and binary activations are not sufficient. Some algorithms are proposed to improve the training algorithm and to increase the accuracy~\cite{Xu21}.

Most of BNNs can be used to handle object detection problems and image classifications problems to achieve acceptable accuracy. However, there are few research papers showing that BNNs can be applied to difficult dense-prediction problems, such as instance segmentation. Bolya et al.~\cite{Bolya19} propose a real-time instance segmentation algorithm, YOLACT, which includes a feature pyramid architecture. Cheng et al.~\cite{Cheng22} propose a fully convolutional framework for real-time instance segmentation, SparseInst, which contains a sparse set of instance activation maps. However, these real-time networks are not quantized and might not be suitable for embedded devices with limited resources because a large number of processing elements are required to calculate the inference results.


\section{Proposed Inference Engine}
\label{sec:proposed}

\begin{table}
\setlength{\tabcolsep}{12pt} 
\caption{Two Operations of BNNs}
\begin{center}
\begin{tabular}{cc|c}
\\
\multicolumn{3}{r}{(a) The First Operation $(m = 0)$}\\
\hline
$a'_{i,j}$ & $w_i$ & $a'_{i,j} \cdot w_i$ \\
\hline
 $0$   &  $-1$  & 0\\
 $0$   &  $+1$  & 0\\
 $+1$   &  $-1$  & $-1$\\
 $+1$   &  $+1$  & $+1$\\
\hline
\\
\multicolumn{3}{r}{(b) The Second Operation $(m = 1)$}\\
\hline
$a'_{i,j}$ & $w'_i$ & $a'_{i,j} \cdot w'_i$ \\
\hline
 $0$    &  $0$  & $0$\\
 $0$    &  $+1$  & $0$\\
 $+1$   &  $0$  & $0$\\
 $+1$   &  $+1$  & $+1$\\
\hline
\end{tabular}
\\
\end{center}
\footnotesize
Note: The values of parameters with the prime symbol ($a'_{i,j}$ and $w'_i$) are $\{0, 1\}$. The values of parameters without the prime symbol ($w_i$) are $\pm 1$.
\label{tab:truthtable}
\end{table}

The proposed inference engine, which can handle BNNs efficiently, is introduced in this section. We focus on the networks where the activations are multi-bit, and the weights are binary. There are $I$ activations, and each activation has $J$ bits, where $J$ is set to $8$ in this work. Each bit of the $i$th activation is represented as $a'_{i,j}$, where $i$ denotes the index of activations, and $j$ denotes the index of bits of the $i$th activation. One weight has only 1 bit, and $w_{i}$ or $w'_{i}$ represents the $i$th weight. Table~\ref{tab:truthtable} shows the two kinds of operations to handle modern BNNs and the corresponding output results. Either of the two operations is selected according to the operation mode $m$, where $m \in \{0, 1\}$. The first mode handles convolutions where $w_{i} \in \{-1, 1\}$, and the second mode handles convolutions or matrix multiplications~\cite{Cheng22} where $w'_{i} \in \{0, 1\}$. Multiply operations can be removed from all MAC operations because only binary weights are employed in the BNNs. The output of MAC operations is shown in Eq.~\ref{eq:conv}.
\begin{eqnarray}
  o = 
  \left\{
  \begin{array}{cl}
  \sum_{i=0}^{I-1} a'_{i} w_{i}  + \beta & \quad \textrm{if } m = 0  \\\\
  \sum_{i=0}^{I-1} a'_{i} w'_{i} + \beta & \quad \textrm{otherwise} 
  \end{array}
  \right.,
\label{eq:conv}
\end{eqnarray}
where $a'_{i, j} \in \{0, 1\}$, $w_{i} \in \{-1, 1\}$, and $w'_{i} \in \{0, 1\}$. 

The operations of the first mode are shown in Table~\ref{tab:truthtable}(a). The concept of XNOR-Net~\cite{Rastegari16} is adopted to handle BNNs with binary weights and binary activations, where both binary weights and binary activations are represented by $\pm 1$. Since there are 3 possible values in the results, some additional operations are included in order to apply the XNOR operations. The output of MAC operations is shown in Eq.~\ref{eq:conv1}.
\begin{eqnarray}
  \sum_{i=0}^{I-1} a'_{i} w_{i} + \beta = \sum_{j=0}^{J-1} \left( 2^{j} \sum_{i=0}^{I-1} a'_{i, j} w_{i} \right) + \beta.
\label{eq:conv1}
\end{eqnarray}
For hardware implementation, the weights are represented as $w'_{i}$, where $w'_{i} \in \{0, 1\}$. Activations have the same values as their original values, but weights do not.

A multi-bit activation $a'_{i}$ can be decomposed into $J$ bits, which can be represented as $\sum_{j=0}^{J-1} 2^{j} \cdot a'_{i,j}$. The variable $\beta$ represents the bias term in the batch normalization~\cite{Yonekawa17} or the quantization process~\cite{Gao19}. The bit-wise operations in the MACs operations can be obtained using Eq.~\ref{eq:bitwise}.
\begin{eqnarray}
  \sum_{i=0}^{I-1} a'_{i, j} w_{i} = \sum_{i=0}^{I-1} \left[ \frac{\left( 2 a'_{i, j} - 1 \right) + 1}{2} \right] w_{i} = 
  \nonumber \\
  \sum_{i=0}^{I-1} (a'_{i, j} \odot w'_{i}) + \frac{\sum_{i=0}^{I-1} w_{i} - I}{2},
\label{eq:bitwise}
\end{eqnarray}
where $\odot$ represents the exclusive-NOR (XNOR) operation, and $a'_{i, j}$ represents the $j$th bit of the $i$th activation. Since $a'_{i, j} \in \{0, 1\}$ and $w'_{i} \in \{0, 1\}$, the XNOR operations can be applied to simplify multiply operations. The output in Eq.~\ref{eq:conv} can be expressed as Eq.~\ref{eq:xnor}.
\begin{eqnarray}
  \sum_{j=0}^{J-1} \left[ 2^{j} \sum_{i=0}^{I-1} (a'_{i, j} \odot w'_{i}) \right] + \nonumber \\
  \sum_{j=0}^{J-1} \left[ 2^{j} \cdot  \frac{\sum_{i=0}^{I-1} \left( w_{i} - 1 \right)}{2}  \right] + \beta = \nonumber \\
  \sum_{i=0}^{I-1} \left[ \sum_{j=0}^{J-1} \left( 2^{j} \cdot a'_{i, j} \odot w'_{i} \right) \right] + \left( \beta + \gamma \right),
\label{eq:xnor}
\end{eqnarray}
where 
\begin{eqnarray}
  \gamma = \left( \frac{\sum_{i=0}^{I-1} w_{i} - I}{2} \right) \cdot (2^{J} - 1).
\label{eq:gamma}
\end{eqnarray}
It is shown that the convolution includes XNOR-accumulate operations and add operations. The variable $\gamma$ is a correction term, which can be computed according to the weights, $w_{i}$ and the number of activations, $I$. Note that $\gamma$ does not require any information related to the activations, and it does not have to be computed during the inference process. The computations of BNNs can be simplified using the quantization algorithms of the IFQ-Net~\cite{Gao19} after substituting the bias term $\beta$ with $\left( \beta + \gamma \right)$.

The operations of the second mode are shown in Table~\ref{tab:truthtable}(b). Different from the first operation, there are only 2 possible values in the results. The output of the MAC operations is shown in Eq.~\ref{eq:conv2}.
\begin{eqnarray}
  \sum_{i=0}^{I-1} a'_{i} w'_{i} + \beta = \sum_{i=0}^{I-1} \left[  \sum_{j=0}^{J-1} \left( 2^{j} \cdot a'_{i, j} \cdot w'_{i} \right) \right] + \beta,
\label{eq:conv2}
\end{eqnarray}
where $a'_{i, j} \in \{0, 1\}$ and $w'_{i} \in \{0, 1\}$. Both activations and weights have the same values as their original values. It is shown that the MAC operations include only AND-accumulate operations and add operations. No correction terms are required in this mode.

Figure~\ref{fig:architecture-a} shows the architecture of the dedicated inference engine, which includes 3 main components and the memory. The main components are the ``bitwise operation and accumulation array," the ``adder array," and the ``quantization and activation unit." The ``bitwise operation and accumulation array" computes the results of Eq.~\ref{eq:conv1} and Eq.~\ref{eq:conv2}. It includes multiple ``bitwise operation units." The architecture is shown in Fig.~\ref{fig:architecture-b}, where each ``bitwise operation unit" includes $J$ ``logical operation units" to calculate the results of XNOR or AND operations. When the first operation mode is enabled, $m$ is set to 0, and the circuit is equivalent to multiple XNOR gates for the input weights and activations. When the second operation mode is enabled, $m$ is set to 1, and the circuit is equivalent to multiple AND gates for the input weights and activations.

The ``adder array" includes parallel adders to compute the sum of the MAC results, the bias term $\beta$, and the correction term $\gamma$. The correction term $\gamma$ does not have to be computed during the inference process. Since it does not require any information related to the activations, it can be computed using any other processors and stored as parameters in advance. This approach can effectively reduce the gate count of the inference engine. Removing the computations only related to weights from embedded devices is one of the key ideas of this work. Besides, the ``adder array" can compute the results of element-wise add operations, which are frequently employed in modern neural networks. The ``quantization and activation unit" quantizes the output of the ``adder array" and computes the pooling results according to the network architecture.

\section{Experimental Results}
\label{sec:experiments}

The experiments are separated into 2 parts, analysis of hardware architectures and BNNs for instance segmentation. The experimental results of both hardware and algorithms are both discussed in this section.

\subsection{BNNs for Instance Segmentation}
\label{tab:exp_algorithm}

\begin{figure*}[t]
\begin{center}
\begin{tabular}{c}
   \includegraphics[width=1.0\linewidth]{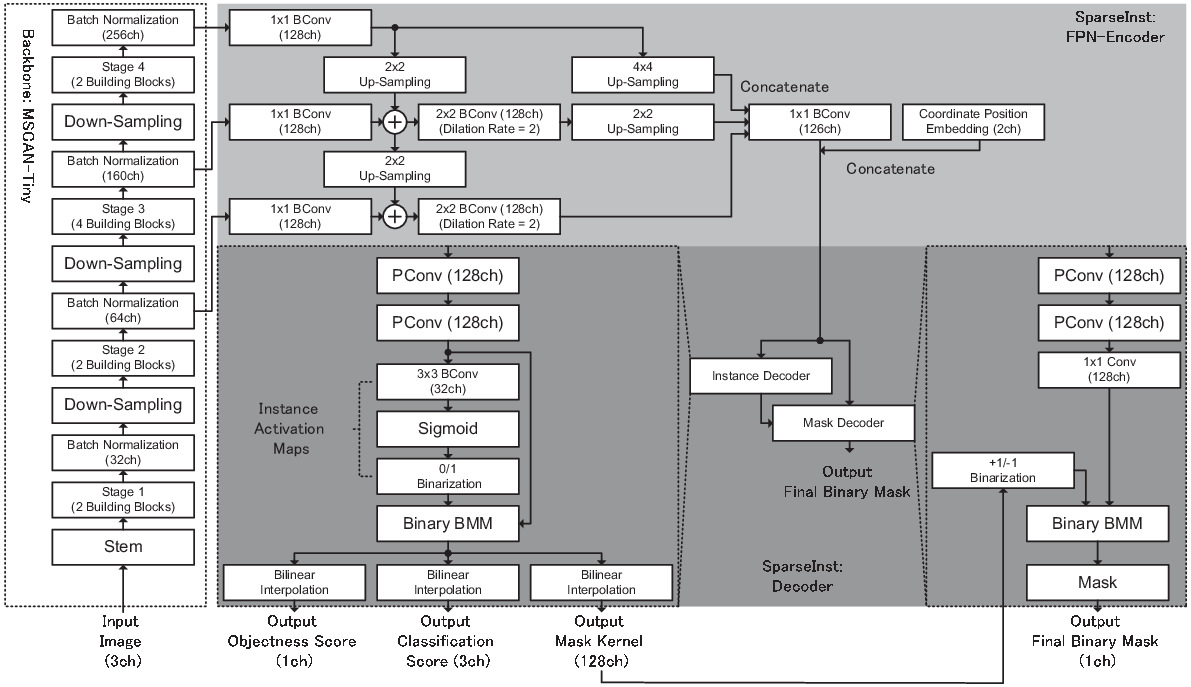}\\
\end{tabular}
\end{center}
   \caption{Architecture of the proposed network, MSCAN-SparseInst BNN.}
\label{fig:networks}
\end{figure*}

\begin{figure}[t]
\begin{center}
\begin{tabular}{c}
   \includegraphics[width=1.0\linewidth]{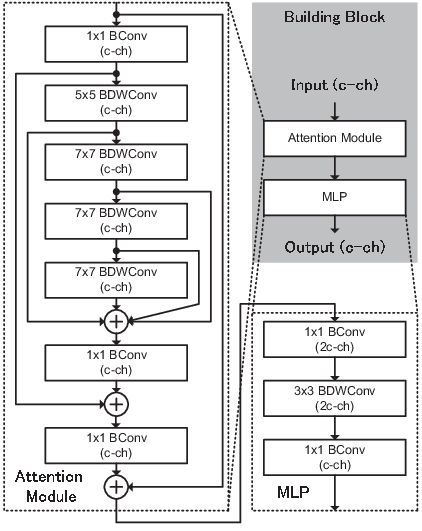}\\
\end{tabular}
\end{center}
   \caption{Architecture of the building block in each stage of MSCAN-SparseInst BNN.}
\label{fig:building}
\end{figure}

\begin{figure*}[t]
\begin{center}
\begin{tabular}{ccc}
   \includegraphics[width=0.31\linewidth]{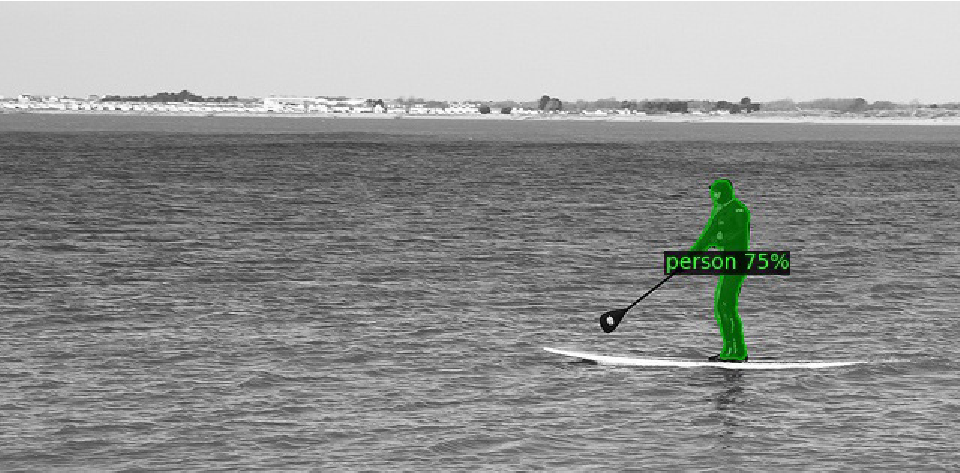} & 
   \includegraphics[width=0.31\linewidth]{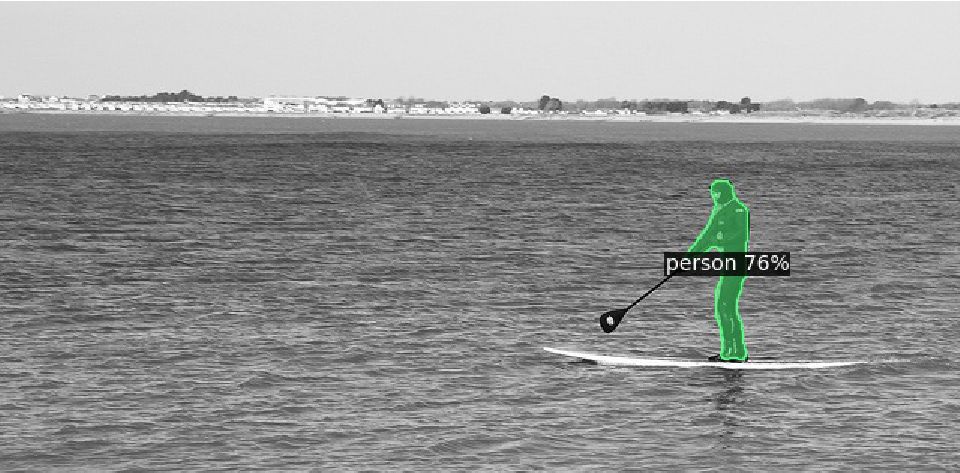} & 
   \includegraphics[width=0.31\linewidth]{000000001490_Yolact.eps}\\
   \includegraphics[width=0.31\linewidth]{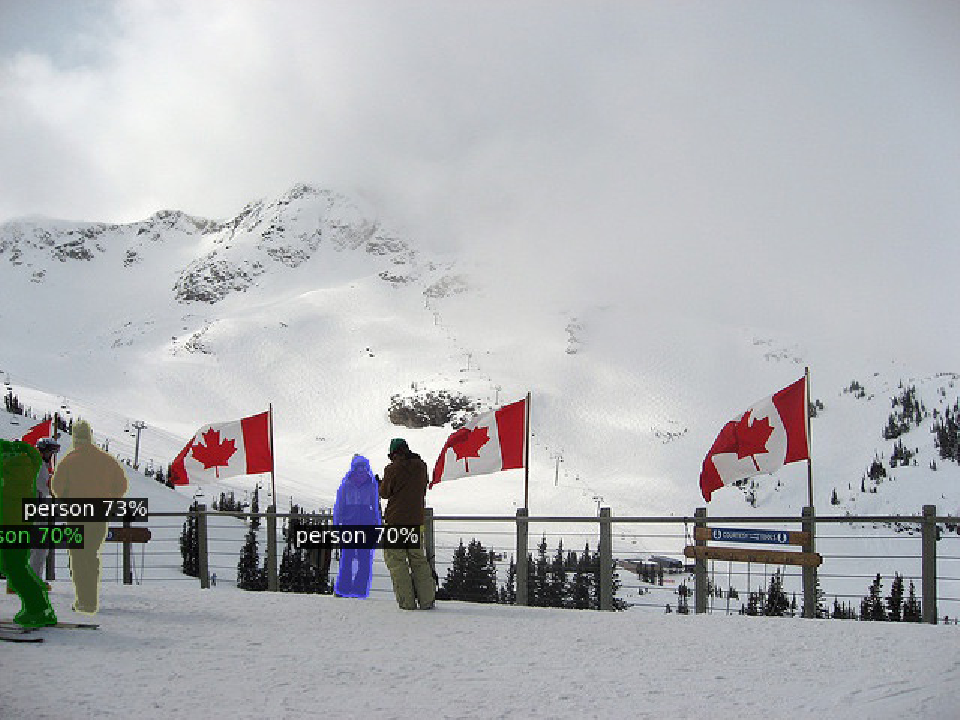} & 
   \includegraphics[width=0.31\linewidth]{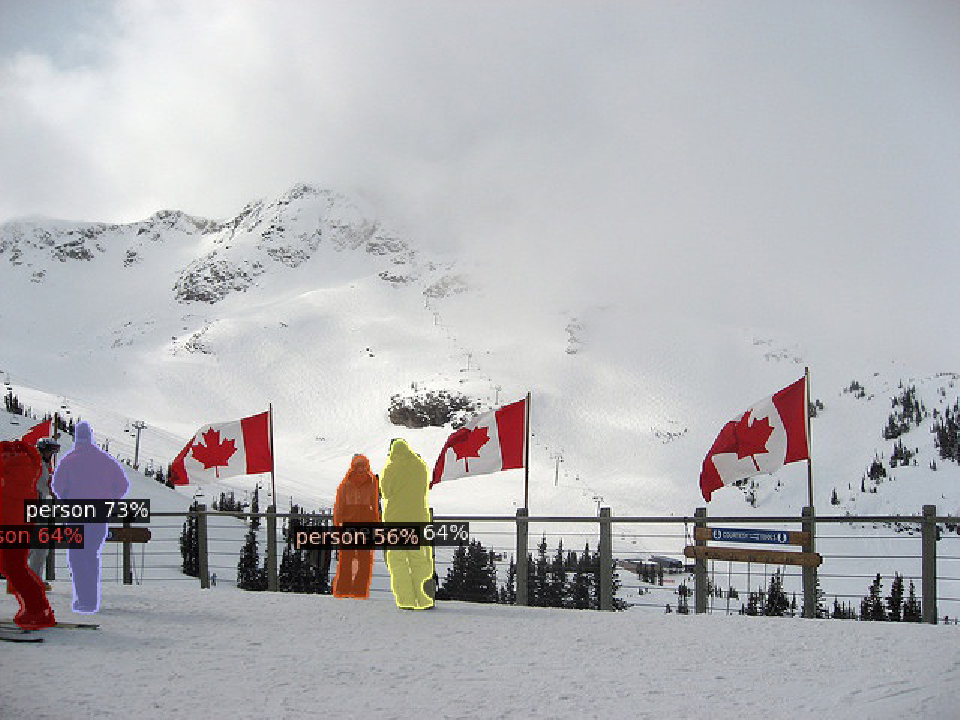} & 
   \includegraphics[width=0.31\linewidth]{000000154087_Yolact.eps}\\   
   \includegraphics[width=0.31\linewidth]{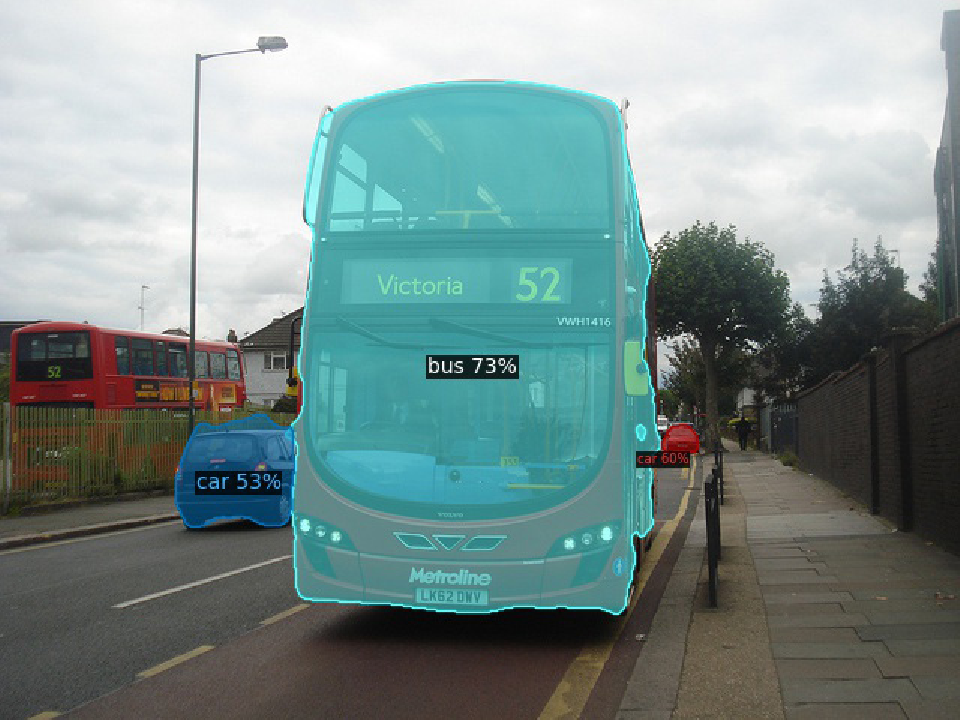} & 
   \includegraphics[width=0.31\linewidth]{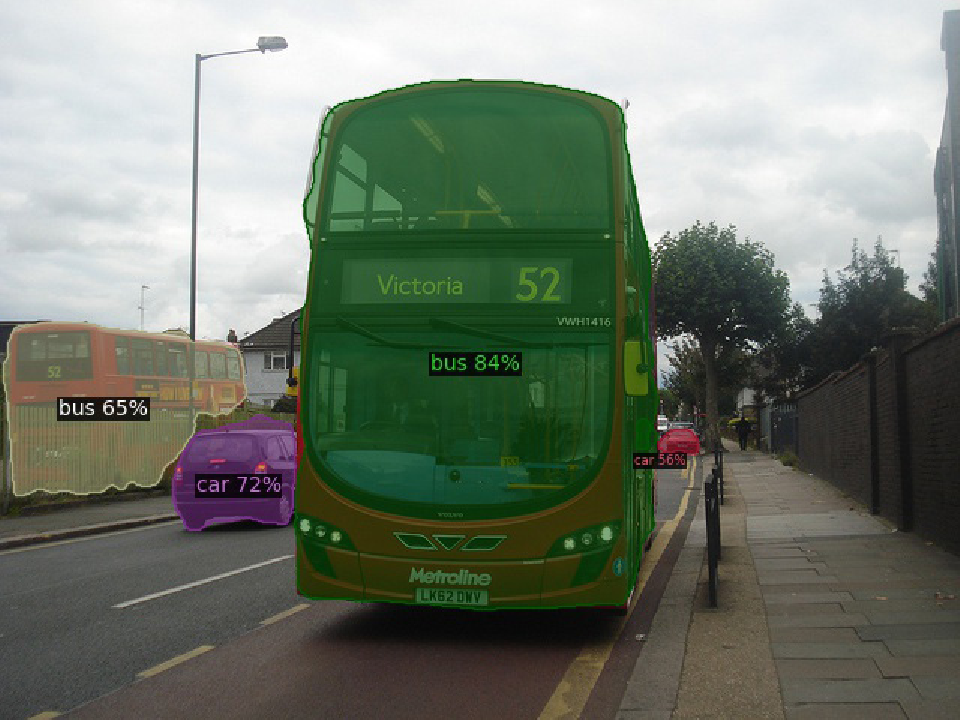} & 
   \includegraphics[width=0.31\linewidth]{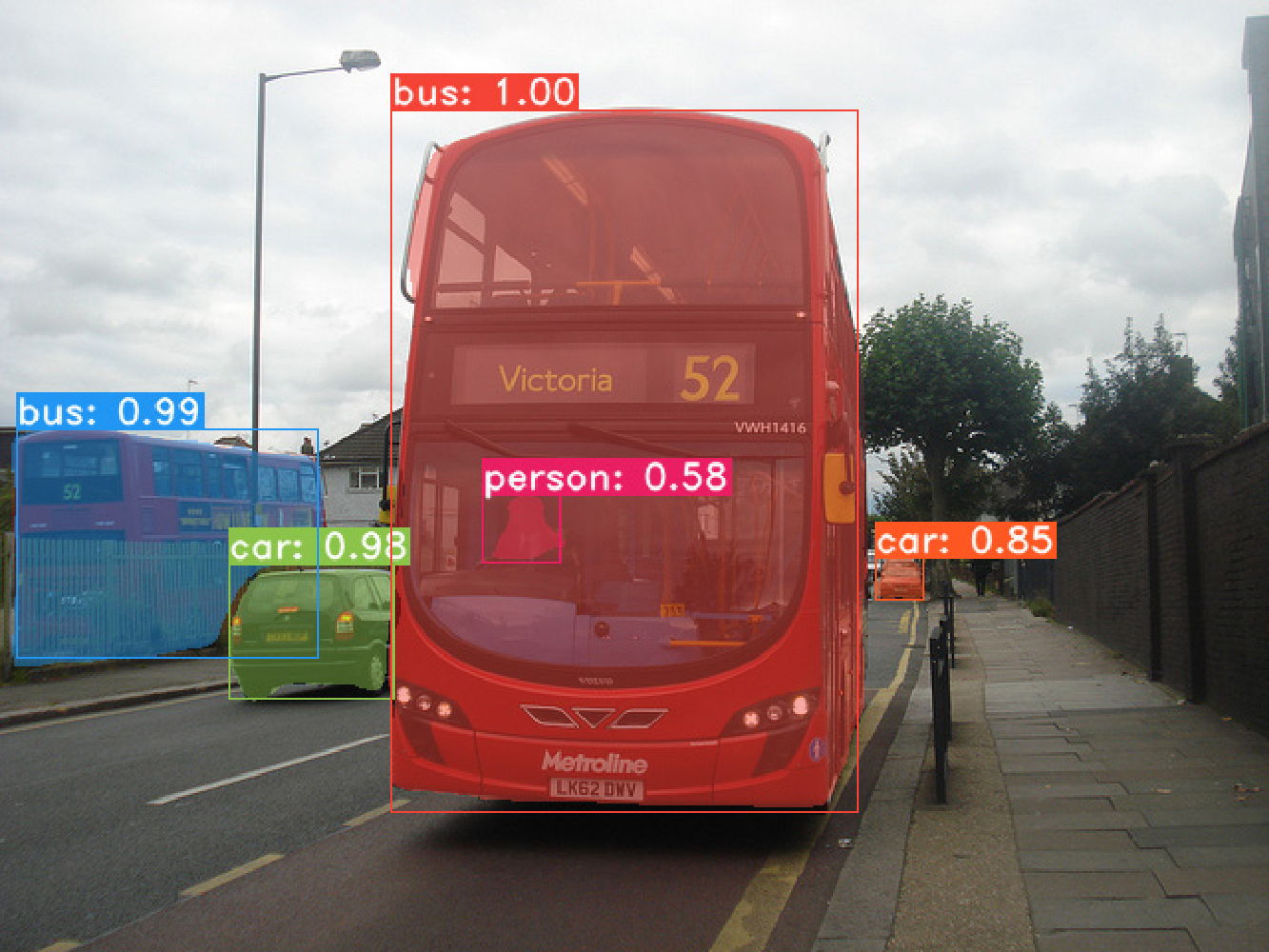}\\
   (a) & (b) & (c) \\
\end{tabular}
\end{center}
   \caption{Some examples of instance segmentation results of (a) ConvNeXtV2-SparseInst BNN, (b) MSCAN-SparseInst BNN, and (c) YOLACT (ResNet50)~\cite{Bolya19}.}
\label{fig:examples}
\end{figure*}

\begin{table*}[t]
\caption{Model Sizes and Computational Costs (VGA Images)}
\setlength{\tabcolsep}{12pt} 
\begin{center}
\begin{tabular}{c|cccc}
\hline
                                  & \multirow{2}{*}{Type}  & Model Size & Computational Cost & \multirow{2}{*}{No. of Layers}\\
                                  &                        & (MB)       & (GMACs)            &\\
\hline                                                             
 YOLACT (ResNet50)~\cite{Bolya19} & Float                  & 34.98$^*$  & 57.4       & 86\\
 MSCAN-SparseInst                 & Float                  & 3.88       & 6.0        & 153\\
 ConvNeXtV2-SparseInst            & Float                  & 3.63       & 5.6        & 139\\
 MSCAN-SparseInst BNN             & A8W1                   & 0.49       & 6.0        & 153\\
 ConvNeXtV2-SparseInst BNN        & A8W1                   & 0.45       & 5.6        & 139\\
\hline
\end{tabular}
\end{center}
\footnotesize
$^{*}$Networks with floating-point weights and activations are used for evaluation, but the model size is evaluated when the bit width of weights is 8 bits.
\label{tab:computations}
\end{table*}

\begin{table*}[t]
\setlength{\tabcolsep}{12pt} 
\caption{Comparison of Accuracy}
\begin{center}
\begin{tabular}{c|cccc}
\hline
                     & \multirow{2}{*}{Type}  & \multicolumn{3}{c}{MAP(\%) (IoU: 0.50 -- 0.95)}\\
\cline{3-5}                         
                     &                       & Person   & Car   & Bus\\
\hline                                               
 YOLACT (ResNet50)~\cite{Bolya19}   & Float    & 27.49  & 25.86 & 59.11\\
 MSCAN-SparseInst                   & Float    & 39.25  & 28.60 & 48.90\\
 ConvNeXtV2-SparseInst              & Float    & 40.22  & 27.87 & 47.96\\
 MSCAN-SparseInst BNN$^1$           & A8W1     & 32.97  & 21.15 & 40.22\\ 
 ConvNeXtV2-SparseInst BNN$^1$      & A8W1     & 31.21  & 21.95 & 39.74\\ 
 MSCAN-SparseInst BNN-A$^2$         & A8W1     & 14.35  & 11.57 & 21.13\\ 
 ConvNeXtV2-SparseInst BNN-A$^2$    & A8W1     & 13.54  & 11.82 & 23.09\\ 
\hline
\end{tabular}
\end{center}
\footnotesize
$^1$With binary matrix multiplications where the values of input data are $\{0,1\}$ and $\pm 1$.\\
$^2$With binary matrix multiplications where the values of input data are $\pm 1$ only.\\
\label{tab:accuracy}
\end{table*}

\begin{table}[t]
\caption{Training Parameters}
\begin{center}
\begin{tabular}{c|c}
\hline                                  
 Dataset                 & COCO 2017 train/val~\cite{Lin14}\\
 Data Augmentation       & Random Flip, crop, resize\\
 Optimization Method     & AdamW$^{*}$\\
 No. of Training Epochs  & 300\\
 Batch Size              & 16\\
 Weight Decay            & 0.05\\
 Multi-scale Training    & Yes\\ 
 \hline
\end{tabular}
\end{center}
\footnotesize
$^{*}$Base learning rate: 0.00005, Betas: 0.9, 0.999.
\label{tab:training}
\end{table}

To evaluate the performance of BNNs, two lightweight networks for instance segmentation, MSCAN-SparseInst BNN and ConvNeXtV2-SparseInst BNN, are designed. Instance segmentation is a representative task of dense prediction, in which pixel-level labeling is required. It is not simple to obtain high accuracy after binarizing the weights or activations in dense-prediction networks. To reduce the computational costs, the decoder of SparseInst~\cite{Cheng22} and the backbone networks of SegNeXt~\cite{Guo22} and ConvNeXtV2~\cite{Woo23} are combined. Besides, some regular convolutions are replaced by the partial convolutions in FasterNet~\cite{Chen23} to reduce computational costs of the proposed networks. The networks are trained using GeForce GTX TITAN X with 12GB memory. Microsoft COCO~\cite{Lin14} is employed for training and evaluation, and the optimization method is AdamW. The techniques in HWGQ~\cite{Cai17}, PACT~\cite{Choi18}, LSQ~\cite{Esser19}, and LSQ+~\cite{Bhalgat20} are used for quantization. The training parameters are shown in Table~\ref{tab:training}. The number of training epochs is 16, and the batch size is 16. To increase the accuracy of BNNs, the training algorithm in PROFIT~\cite{Park20}, which progressively freezes the most sensitive layer of the network, is also employed.

The architecture of one of the proposed networks, MSCAN-SparseInst BNN, is shown in Fig.~\ref{fig:networks}. BConv denotes the convolutions with binary weights, and PConv represents partial convolutions in FasterNet~\cite{Chen23}. The backbone network is modified from MSCAN-Tiny, which is included in the architecture of SegNeXt~\cite{Guo22}. It contains 3 down-sampling layers and 4 stages, which are followed by 4 batch normalization (BN) layers, and the number of building blocks in each stage is different. The architecture of the building block in each stage in shown in Fig.~\ref{fig:building}, where BDWConv denotes the depthwise convolutions with binary weights. The rectangular filters in the original MSCAN are removed to simplify the network. The activations from 3 BN layers are sent to the FPN-Encoder, which is modified from the architecture of SparseInst~\cite{Cheng22}. The FPN-Encoder contains 4 nearest-neighbor up-sampling layers, 6 convolution layers, and 1 coordinate position embedding layer. The input of the coordinate position embedding layer is 126, and the output is 128 channels since 2 coordinate channels (x and y) are added into the layer. The activations from the coordinate position embedding layer are sent to the decoder, which is modified from the architecture of SparseInst~\cite{Cheng22}. The decoder contains 6 convolution layers, and 2 Batch-Matrix-Matrix (BMM) layers. The inputs of 2 BMM layers are binarized to $\{0,1\}$ and $\pm 1$, respectively. To compare different network architectures, a network architecture modified from ConvNeXtV2-femto, which is the backbone network in ConvNeXtV2~\cite{Woo23}, is also evaluated.

In the two lightweight networks, the weights in the convolutions layers are all quantized to $\pm 1$, and the first mode ($m = 0$) of the inference engine can be applied. In the decoder of the networks, there are some matrix operations in BMM layers~\cite{Cheng22} employed to increase the accuracy. One matrix multiplication has 2 input activations. One of them is quantized to $\{0,1\}$ or $\pm 1$, so that the MAC operations can be handled with the two modes ($m = 0$ or $m = 1$) of the proposed hardware. The proposed networks, ConvNeXtV2-SparseInst and MSCAN-SparseInst, are compared with YOLACT~\cite{Bolya19}. The model sizes and the computational costs are shown in Table~\ref{tab:computations}. The full-precision (floating-point) version of ConvNeXtV2-SparseInst is 9.6$\times$ smaller than YOLACT when the bit width of weights is 8 bits, and the binary version of ConvNeXtV2-SparseInst (ConvNeXtV2-SparseInst BNN) is 77.7$\times$ smaller than YOLACT because the weights are reduced from 8 bits to 1 bit. Compared with YOLACT, ConvNeXtV2-SparseInst BNN has only 9.8\% of MACs, and the weights are binarized. The results of MSCAN-SparseInst and MSCAN-SparseInst BNN are also similar. The computational costs and the model sizes of the two networks are effectively reduced compared with YOLACT although they have more layers than YOLACT.

The accuracy is shown in Table~\ref{tab:accuracy}. The mean average precision (MAP) of each of the three categories, ``Person," ``Car," and ``Bus," is evaluated. Some examples of instance segmentation results are shown in Fig.~\ref{fig:examples}, where it is observed that the two lightweight networks have higher detection rates on the ``Person" category than YOLACT~\cite{Bolya19}. Two different versions of binary networks are designed for the two proposed networks. The BMM layer in the decoder of SparseInst~\cite{Cheng22} includes two floating-point input activations. In order to make them compatible with the proposed hardware, one of the input activations is quantized to 1 bit. Also, the normalization operations in the Instance Activation Maps (IAM) of SparseInst~\cite{Cheng22} are removed to simplify the operations. The results show that both MSCAN-SparseInst BNN and ConvNeXtV2-SparseInst achieve higher accuracy than YOLACT on the ``Person" category after binarizing the input of BMM operations. Moreover, MSCAN-SparseInst BNN-A and ConvNeXtV2-SparseInst BNN-A are designed for ablation study. MSCAN-SparseInst BNN-A and MSCAN-SparseInst BNN have the same network architecture, but the values of the inputs to the BMM layers in MSCAN-SparseInst BNN-A are quantized to only $\pm 1$, not $\{0,1\}$ and $\pm 1$. MSCAN-SparseInst BNN has higher accuracy on the ``Person" category (+18.6\%) than MSCAN-SparseInst BNN-A. It means that the functions to support two kinds of binary weights effectively increase the accuracy of instance segmentation. The results of ConvNeXtV2-SparseInst BNN also show the importance of the input values of binary operations.

\subsection{Analysis of Hardware Architectures}
\label{tab:exp_hardware}

\begin{figure}[t]
\begin{center}
\begin{tabular}{cc}
   \includegraphics[width=0.8\linewidth]{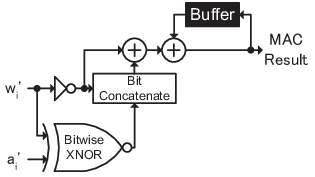}\\
   (a)\\
   \includegraphics[width=0.8\linewidth]{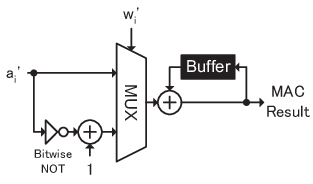}\\
   (b)
\end{tabular}
\end{center}
   \caption{Hardware architecture of (a) the XNOR-based multiplier~\cite{Guo19,Yonekawa17} with additional logics, and (b) the selector-based multiplier~\cite{Guo23}.}
\label{fig:related}
\end{figure}

\begin{figure}[t]
\begin{center}
\begin{tabular}{cc}
   \includegraphics[width=0.9\linewidth]{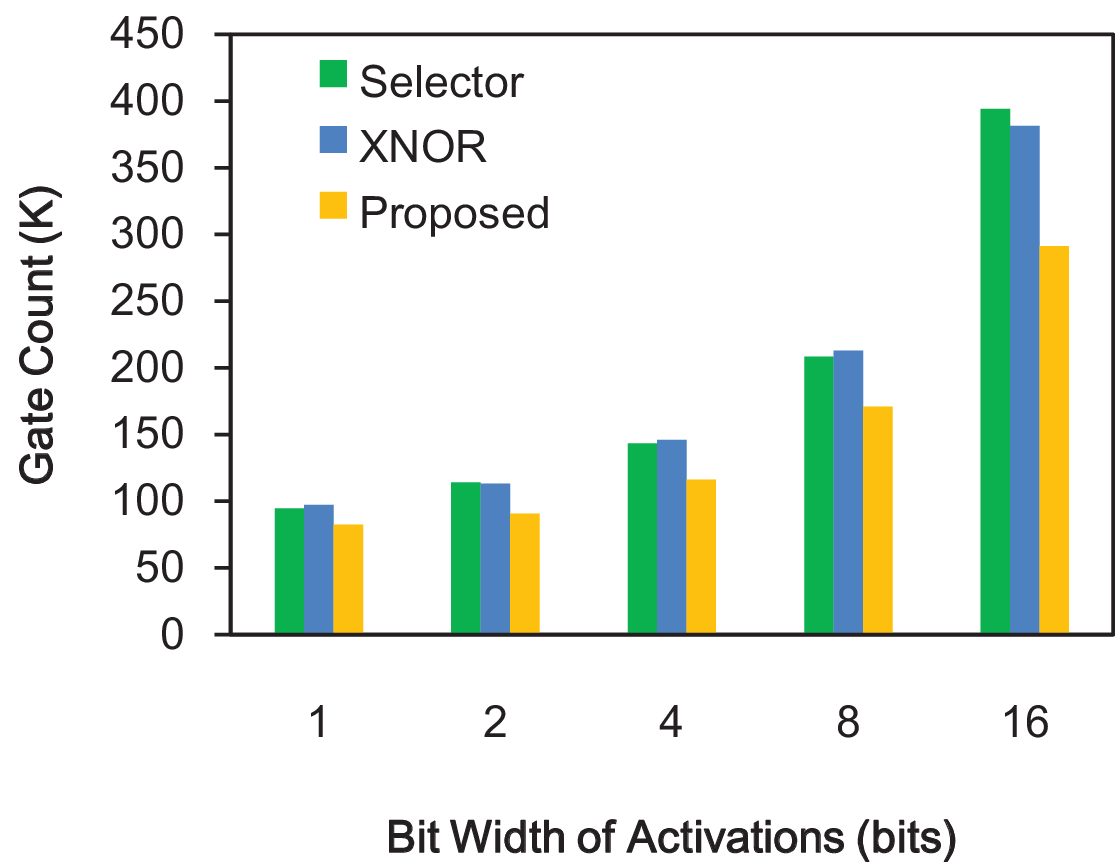}\\ 
   (a)\\
   \includegraphics[width=0.9\linewidth]{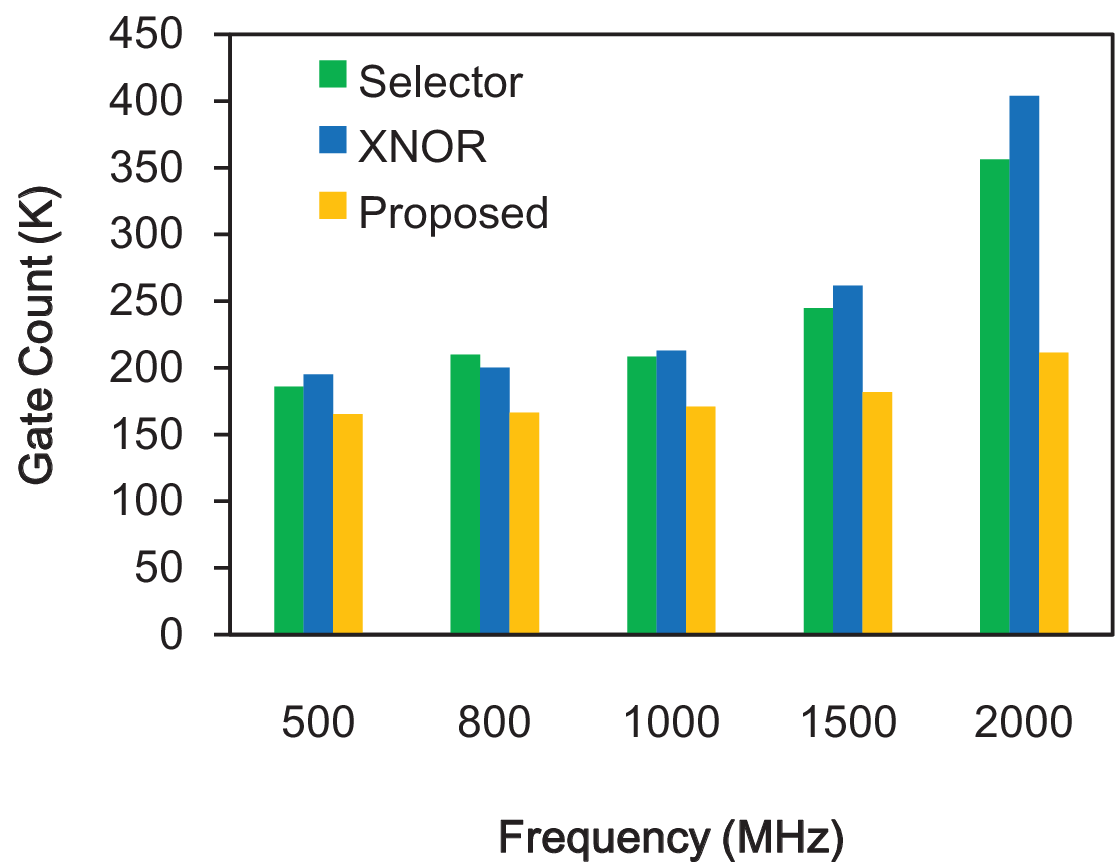}\\
   (b)
\end{tabular}
\end{center}
   \caption{Comparison of gate counts among the selector-based multiplier \cite{Guo23}, the XNOR-based multiplier~\cite{Guo19,Yonekawa17} with additional logics, and the ``bitwise operation and accumulation array" in the proposed inference engine. (a) Operating frequency is set to 1GHz. (b) Bit width of activations is set to 8 bits.}
\label{fig:analysis}
\end{figure}

\begin{table}[t]
\caption{Comparison of Implementation Methods}
\setlength{\tabcolsep}{12pt} 
\begin{center}
\begin{tabular}{c|c}
\hline
                                         & Supported Network \\
 \hline                                                 
 XNOR operations~\cite{Guo19,Yonekawa17} & A1W1         \\
 XOR operations~\cite{Zhu20}             & A1W1         \\
 Select operations~\cite{Guo23}          & A$J$W1$^{*}$ \\
\hline
\end{tabular}
\end{center}
\footnotesize
$^*$Activations are quantized to $J$ bits, and weights are quantized to 1 bit.\\
\label{tab:comparison1}
\end{table}

\begin{table*}[t]
\caption{Comparison of Different Multipliers for 1-Bit Weights (Operating Frequency: 2GHz)}
\setlength{\tabcolsep}{12pt} 
\begin{center}
\begin{tabular}{c|ccc}
\hline
                                        & Supported Network   & Binary Weights               & Gate Count\\
\hline                                                 
 Selector-based Multiplier~\cite{Guo23} & A8W1       & \multirow{2}{*}{$\pm 1$}  & 356K\\ 
 Modified XNOR-based Multiplier$^*$     & A8W1       &                           & 404K\\
 This Work                              & A8W1       & $\{0,1\}$ or $\pm 1$       & 211K\\
\hline
\end{tabular}
\end{center}
\footnotesize
$^*$Modified XNOR-based multiplier~\cite{Guo19,Yonekawa17} with additional logics.\\
\label{tab:comparison2}
\end{table*}

The proposed MAC architecture (the ``bitwise operation and accumulation array") is compared with related works, which are shown in Table~\ref{tab:comparison1}. The values of binary weights supported by the related works~\cite{Guo23,Guo19,Yonekawa17,Zhu20} are $\pm 1$. The architectures implemented with XNOR~\cite{Guo19,Yonekawa17} and XOR~\cite{Zhu20} operations are designed to handle 1-bit weights and 1-bit activations. To compare them with this work, some additional logic operations and adders are added into the XNOR-based multiplier~\cite{Guo19,Yonekawa17} to support multi-bit weights. The modified architecture is shown in Fig.~\ref{fig:related}(a). The hardware architecture implemented with select operations~\cite{Guo23} is designed to handle multi-bit weights and 1-bit activations. The architecture, which is shown in Fig.~\ref{fig:related}(b), can be directly compared with this work. The proposed work is designed to handle multi-bit activations and 1-bit weights, and the values of binary weights can be either $\{0,1\}$ or $\pm 1$. 

The relation between the gate count of the architectures and the bit width of activations is shown in Fig.~\ref{fig:analysis}. The modules of the related works are re-implemented to fit the proposed hardware architecture and synthesized with the regular threshold voltage (RVT) device model in the ASAP7 library \cite{Vashishtha17}. The results in Fig.~\ref{fig:analysis}(a) show that the modified XNOR-based multiplier has similar gate counts with the selector-based multiplier~\cite{Guo23}, and the trend does not change with the bit width of activations. The results in Fig.~\ref{fig:analysis}(b) show that, compared with the related works, the gate count of this work is reduced to 52\% of the modified XNOR-based multiplier and 59\% of the selector-based multiplier~\cite{Guo23} when the bit width of activations is 8 and the operating frequency is 2GHz. The higher the operating frequency, the more reduction of gate counts. The comparison of the related works and this work is summarized in Table~\ref{tab:comparison2}. The proposed work has lower costs than the related works because the computations of the correction term shown in Eq.~\ref{eq:gamma} are removed from the MAC operations. Since the correction term is merged into the bias term, no additional operations are required in the inference process. Moreover, this work can support two kinds of binary weights, $\{0,1\}$ and $\pm 1$, which increase the accuracy of instance segmentation as shown in Sec.~\ref{tab:exp_algorithm}.

\section{Conclusion}
\label{sec:conclusions}
A hardware architecture and a design methodology of dedicated inference engines for BNNs are proposed. The proposed inference engine can handle instance segmentation with only bitwise operations and add operations. The architecture of MAC operations can calculate the inference results of BNNs efficiently with only 52\% of hardware costs compared with the related works. A part of computation costs can be removed from the system because they are not dependent on activation maps and can be performed in advance using any other processors. 

In addition, two lightweight networks for instance segmentation and hardware architecture of inference engine are proposed. The experimental results show that the proposed inference engine can handle the proposed instance-segmentation networks and achieves higher accuracy as YOLACT on the ``Person" category although the model size is 77.7$\times$ smaller compared with YOLACT.

\section*{Acknowledgment}
This paper is based on results obtained from a project, JPNP23015, commissioned by the New Energy and Industrial Technology Development Organization (NEDO).

{\small
\bibliographystyle{ieee_fullname}
\bibliography{IEEEfull,cvprws2024_paper}

\begin{thebibliography}{10}\itemsep=-1pt

\bibitem{Agrawal21}
Ankur Agrawal, Sae~Kyu Lee, Joel Silberman, Matthew Ziegler, Mingu Kang,
  Swagath Venkataramani, Nianzheng Cao, Bruce Fleischer, Michael Guillorn,
  Matthew Cohen, Silvia Mueller, Jinwook Oh, Martin Lutz, Jinwook Jung, Siyu
  Koswatta, Ching Zhou, Vidhi Zalani, James Bonanno, Robert Casatuta, Chia-Yu
  Chen, Jungwook Choi, Howard Haynie, Alyssa Herbert, Radhika Jain, Monodeep
  Kar, Kyu-Hyoun Kim, Yulong Li, Zhibin Ren, Scot Rider, Marcel Schaal, Kerstin
  Schelm, Michael Scheuermann, Xiao Sun, Hung Tran, Naigang Wang, Wei Wang, Xin
  Zhang, Vinay Shah, Brian Curran, Vijayalakshmi Srinivasan, Pong-Fei Lu, Sunil
  Shukla, Leland Chang, and Kailash Gopalakrishnan.
\newblock A 7nm 4-core {AI} chip with {25.6TFLOPS} hybrid {FP8} training,
  {102.4TOPS} {INT4} inference and workload-aware throttling.
\newblock In {\em Proceedings of IEEE International Solid-State Circuits
  Conference}, pages 144--146, Feb. 2021.

\bibitem{Bhalgat20}
Yash Bhalgat, Jinwon Lee, Markus Nagel, Tijmen Blankevoort, and Nojun Kwak.
\newblock {LSQ}+: Improving low-bit quantization through learnable offsets and
  better initialization, 2020.
\newblock CoRR, abs/2004.09576.

\bibitem{Bolya19}
Daniel Bolya, Chong Zhou, Fanyi Xiao, and Yong~Jae Lee.
\newblock {YOLACT:} real-time instance segmentation, 2019.
\newblock CoRR, abs/1904.02689.

\bibitem{Cai17}
Zhaowei Cai, Xiaodong He, Jian Sun, and Nuno Vasconcelos.
\newblock Deep learning with low precision by half-wave {Gaussian}
  quantization, 2017.
\newblock CoRR, abs/1702.00953.

\bibitem{Chen23}
Jierun Chen, {Shiu-hong} Kao, Hao He, Weipeng Zhuo, Song Wen, Chul-Ho Lee, and
  S.-H.~Gary Chan.
\newblock Run, don’t walk: Chasing higher {FLOPS} for faster neural networks,
  2023.
\newblock CoRR, abs/2303.03667v3.

\bibitem{Chen21}
Tse-Wei Chen, Motoki Yoshinaga, Hongxing Gao, Wei Tao, Dongchao Wen, Junjie
  Liu, Kinya Osa, and Masami Kato.
\newblock {Condensation-Net:} memory-efficient network architecture with
  cross-channel pooling layers and virtual feature maps, 2021.
\newblock CoRR, abs/2104.14124.

\bibitem{Cheng22}
Tianheng Cheng, Xinggang Wang, Shaoyu Chen, Wenqiang Zhang, Qian Zhang, Chang
  Huang, Zhaoxiang Zhang, and Wenyu Liu.
\newblock Sparse instance activation for real-time instance segmentation, 2022.
\newblock CoRR, abs/2203.12827v1.

\bibitem{Choi18}
Jungwook Choi, Zhuo Wang, Swagath Venkataramani, Pierce I-Jen Chuang,
  Vijayalakshmi Srinivasan, and Kailash Gopalakrishnan.
\newblock {PACT:} parameterized clipping activation for quantized neural
  networks, 2018.
\newblock CoRR, abs/1805.06085v2.

\bibitem{Chung23}
Ching-Che Chung, Yu-Pei Liang, Ya-Ching Chang, and Chen-Ming Chang.
\newblock A binary weight convolutional neural network hardware accelerator for
  analysis faults of the {CNC} machinery on {FPGA}.
\newblock In {\em Proceedings of International VLSI Symposium on Technology,
  Systems and Applications (VLSI-TSA/VLSI-DAT)}, Apr. 2023.

\bibitem{Courbariaux15}
Matthieu Courbariaux, Yoshua Bengio, and Jean-Pierre David.
\newblock {BinaryConnect:} training deep neural networks with binary weights
  during propagations, 2015.
\newblock CoRR, abs/1511.00363.

\bibitem{Csurka23}
Gabriela Csurka, Riccardo Volpi, and Boris Chidlovskii.
\newblock Semantic image segmentation: Two decades of research, 2023.
\newblock CoRR, abs/2302.06378.

\bibitem{Deng19}
Jiankang Deng, Jia Guo, Yuxiang Zhou, Jinke Yu, Irene Kotsia, and Stefanos
  Zafeiriou.
\newblock {RetinaFace}: Single-stage dense face localisation in the wild, 2019.
\newblock CoRR, abs/1905.00641.

\bibitem{Esser19}
Steven~K. Esser, Jeffrey~L. McKinstry, Deepika Bablani, Rathinakumar Appuswamy,
  and Dharmendra~S. Modha.
\newblock Learned step size quantization, 2019.
\newblock CoRR, abs/1902.08153.

\bibitem{Gao19}
Hongxing Gao, Wei Tao, Dongchao Wen, Tse-Wei Chen, Kinya Osa, and Masami Kato.
\newblock {IFQ-Net:} integrated fixed-point quantization networks for embedded
  vision, 2019.
\newblock CoRR, abs/1911.08076.

\bibitem{Guo23}
Lunyi Guo, Shining Mu, Yijie Deng, Chaofan Shi, Bo Yan, and Zhuoling Xiao.
\newblock Efficient binary weight convolutional network accelerator for speech
  recognition.
\newblock {\em Sensors}, 23(3):1530, May 2023.

\bibitem{Guo22}
Meng-Hao Guo, Cheng-Ze Lu, Qibin Hou, Zhengning Liu, Ming-Ming Cheng, and
  Shi-Min Hu.
\newblock {SegNeXt:} rethinking convolutional attention design for semantic
  segmentation, 2022.
\newblock CoRR, abs/2209.08575.

\bibitem{Guo19}
Peng Guo, Hong Ma, Ruizhi Chen, and Donglin Wang.
\newblock A high-efficiency {FPGA}-based accelerator for binarized neural
  network.
\newblock {\em Journal of Circuits, Systems and Computers}, 28(supp01), Apr.
  2019.

\bibitem{Hu18}
Yuwei Hu, Jidong Zhai, Dinghua Li, Yifan Gong, Yuhao Zhu, Wei Liu, Lei Su, and
  Jiangming Jin.
\newblock {BitFlow}: Exploiting vector parallelism for binary neural networks
  on {CPU}.
\newblock In {\em Proceedings of International Parallel and Distributed
  Processing Symposium}, pages 244--253, May 2018.

\bibitem{Sambhav20}
Sambhav~R. Jain, Albert Gural, Michael Wu, and Chris~H. Dick.
\newblock Trained quantization thresholds for accurate and efficient
  fixed-point inference of deep neural networks, 2020.
\newblock CoRR, abs/1903.08066v3.

\bibitem{Lin14}
Tsung-Yi Lin, Michael Maire, Serge Belongie, Lubomir Bourdev, Ross Girshick,
  James Hays, Pietro Perona, Deva Ramanan, C.~Lawrence Zitnick, and Piotr
  Dollár.
\newblock {Microsoft COCO:} common objects in context, 2014.
\newblock CoRR, abs/1405.0312.

\bibitem{Park20}
Eunhyeok Park and Sungjoo Yoo.
\newblock {PROFIT:} a novel training method for sub-4-bit {MobileNet} models,
  2020.
\newblock CoRR, abs/2008.04693.

\bibitem{Qin20}
Haotong Qin, Ruihao Gong, Xianglong Liu, Xiao Bai, Jingkuan Song, and Nicu
  Sebe.
\newblock Binary neural networks: A survey, 2020.
\newblock CoRR, abs/2004.03333.

\bibitem{Ranftl21}
Ren{\"e} Ranftl, Alexey Bochkovskiy, and Vladlen Koltun.
\newblock Vision transformers for dense prediction, 2021.
\newblock CoRR, abs/2103.13413v1.

\bibitem{Rastegari16}
Mohammad Rastegari, Vicente Ordonez, Joseph Redmon, and Ali Farhadi.
\newblock {XNOR-Net: ImageNet} classification using binary convolutional neural
  networks, 2016.
\newblock CoRR, abs/1603.05279v4.

\bibitem{Terven23}
Juan Terven and Diana Cordova-Esparza.
\newblock A comprehensive review of {YOLO} architectures in computer vision:
  From {YOLOv1 to YOLOv8 and YOLO-NAS}, 2024.
\newblock CoRR, abs/2304.00501v7.

\bibitem{Tsai20}
Tsung-Han Tsai and Yuan-Chen Ho.
\newblock A {CNN} accelerator on {FPGA} using binary weight networks.
\newblock In {\em Proceedings of IEEE International Conference on Consumer
  Electronics (ICCE)}, pages 1--2, Sept. 2020.

\bibitem{Vashishtha17}
Vinay Vashishtha, Manoj Vangala, and Lawrence~T. Clark.
\newblock {ASAP7} predictive design kit development and cell design technology
  co-optimization: Invited paper.
\newblock In {\em Proceedings of The International Conference on Computer-Aided
  Design (ICCAD)}, pages 992--998, Nov. 2017.

\bibitem{Woo23}
Sanghyun Woo, Shoubhik Debnath, Ronghang Hu, Xinlei Chen, Zhuang Liu, In~So
  Kweon, and Saining Xie.
\newblock {ConvNeXt V2:} co-designing and scaling {ConvNets} with masked
  autoencoders, 2023.
\newblock CoRR, abs/2301.00808.

\bibitem{Xu21}
Zihan Xu, Mingbao Lin, Jianzhuang Liu, Jie Chen, Ling Shao, Yue Gao, Yonghong
  Tian, and Rongrong Ji.
\newblock {ReCU:} reviving the dead weights in binary neural networks, 2021.
\newblock CoRR, abs/2103.12369v2.

\bibitem{Yonekawa17}
Haruyoshi Yonekawa and Hiroki Nakahara.
\newblock On-chip memory based binarized convolutional deep neural network
  applying batch normalization free technique on an {FPGA}.
\newblock In {\em Proceedings of IEEE International Parallel and Distributed
  Processing Symposium Workshops}, pages 98--105, May 2017.

\bibitem{Zhu20}
Shien Zhu, Luan H.~K. Duong, and Weichen Liu.
\newblock {XOR-Net}: An efficient computation pipeline for binary neural
  network inference on edge devices.
\newblock In {\em Proceedings of International Conference on Parallel and
  Distributed Systems (ICPADS)}, pages 124--131, Feb. 2020.

\end{thebibliography}
}
\end{document}